\documentclass{article}

\usepackage{arxiv}

\usepackage[utf8]{inputenc} 
\usepackage[T1]{fontenc}    
\usepackage{hyperref}       
\usepackage{url}            
\usepackage{booktabs}       
\usepackage{amsfonts}       
\usepackage{nicefrac}       
\usepackage{microtype}      
\usepackage{lipsum}		
\usepackage{graphicx}
\usepackage{natbib}
\usepackage{doi}

\usepackage{times}
\usepackage{amsfonts}
\usepackage{amssymb}
\usepackage{caption}                 
\usepackage{lipsum}                  
\usepackage{soul}
\usepackage{url}
\usepackage[utf8]{inputenc}

\usepackage{amsmath}
\usepackage{graphicx}   
\usepackage{caption}    
\usepackage{amsmath}    
\usepackage{amsthm}
\usepackage{booktabs}
\usepackage{algorithm}
\usepackage[switch]{lineno}

\usepackage{url}
\usepackage{dsfont}
\usepackage{graphicx}
\usepackage{multirow}
\usepackage{booktabs}
\usepackage{amsmath}
\usepackage{amssymb}
\usepackage{xcolor}
\usepackage[table]{xcolor}
\usepackage{pifont}
\usepackage{caption}
\usepackage{subcaption}
\usepackage{wrapfig}
\usepackage{algorithm}
\usepackage{algpseudocode}
\usepackage{natbib}

\definecolor{purple}{rgb}{0.56,0.27,0.68}
\definecolor{newred}{rgb}{0.95,0.4,0.4}
\definecolor{purered}{rgb}{1,0,0}
\definecolor{blue}{rgb}{0.4,0.4,0.95}
\definecolor{darkblue}{rgb}{0,0,0.8}
\definecolor{grey}{rgb}{0.6,0.6,0.6}
\definecolor{col1}{RGB}{232, 161, 148}
\definecolor{col2}{RGB}{148, 187, 232}
\definecolor{col3}{RGB}{206, 239, 255}
\definecolor{lightgrey}{rgb}{0.85,0.85,0.85}
\definecolor{lightlightgrey}{rgb}{0.9,0.9,0.9}
\definecolor{verylightBG}{rgb}{0.9,0.99,0.99}
\definecolor{darkgreen}{rgb}{0.3, 0.75, 0.3}
\definecolor{orange}{rgb}{1.0,0.65,0.1}
\definecolor{darkorange}{rgb}{1.0,0.549,0.0}
\definecolor{cvprblue}{rgb}{0.21,0.49,0.74}

\hypersetup{
    colorlinks=true,
    linkcolor=darkorange,
    citecolor=cvprblue,
    urlcolor=magenta
}

\title{Virtual Biopsy for Intracranial Tumors Diagnosis on MRI}

\author{
  \href{https://orcid.org/0009-0003-7771-4765}{\includegraphics[scale=0.06]{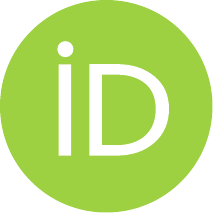}\hspace{1mm}Xinzhe Luo} \quad
  {\includegraphics[scale=0.06]{orcid.pdf}\hspace{1mm}Shuai Shao} \\
  School of Artificial Intelligence and Data Science\\
  University of Science and Technology of China\\
  \AND
  {\includegraphics[scale=0.06]{orcid.pdf}\hspace{1mm}Yan Wang} \quad
  {\includegraphics[scale=0.06]{orcid.pdf}\hspace{1mm}Jiangtao Wang} \quad
  {\includegraphics[scale=0.06]{orcid.pdf}\hspace{1mm}Yutong Bai} \quad
  {\includegraphics[scale=0.06]{orcid.pdf}\hspace{1mm}Jianguo Zhang}
}

\renewcommand{\shorttitle}{\textit{arXiv} Template}

\hypersetup{
pdftitle={A template for the arxiv style},
pdfsubject={q-bio.NC, q-bio.QM},
pdfauthor={David S.~Hippocampus, Elias D.~Striatum},
pdfkeywords={First keyword, Second keyword, More},
}
\begin{document}
\maketitle

\begin{abstract}
	Deep intracranial tumors situated in eloquent brain regions controlling vital functions present critical diagnostic challenges. Clinical practice has shifted toward stereotactic biopsy for pathological confirmation before treatment. Yet biopsy carries inherent risks of hemorrhage and neurological deficits, and struggles with sampling bias due to tumor spatial heterogeneity, because pathological changes are typically region-selective rather than tumor-wide. Therefore, advancing non-invasive MRI-based pathology prediction is essential for holistic tumor assessment and modern clinical decision-making. The primary challenge lies in data scarcity: low tumor incidence requires long collection cycles, and annotation demands biopsy-verified pathology from neurosurgical experts. Additionally, tiny lesion volumes lacking segmentation masks cause critical features to be overwhelmed by background noise. To address these challenges, we construct the ICT-MRI dataset—the first public biopsy-verified benchmark with 249 cases across four categories. We propose a Virtual Biopsy framework comprising: MRI-Processor for standardization, Tumor-Localizer employing vision-language models for coarse-to-fine localization via weak supervision, and Adaptive-Diagnoser with Masked Channel Attention mechanism fusing local discriminative features with global contexts. Experiments demonstrate over 90\% accuracy, outperforming baselines by more than 20\%.
\end{abstract}


\section{Introduction}

\textbf{Background.} 
With the rising incidence of malignant brain tumors, formulating scientific and effective treatment strategies is paramount for maximizing tumor control and prolonging patient survival~\citep{khalighi2024artificial}. Clinical management relies heavily on tumor location. For accessible tumors in non-functional areas, direct surgical resection remains the standard of care.
Conversely, \textit{for tumors situated in deep, eloquent regions (i.e.,  \textbf{Intracranial Tumors)}, surgical resection is fraught with extreme risk.} These areas are densely populated with critical neural nuclei controlling respiration, heartbeat, and motor functions. Furthermore, such tumors often exhibit infiltrative growth with indistinct boundaries, rendering precise dissection without functional damage nearly impossible. Aggressive resection attempts in this context can precipitate catastrophic outcomes, including irreversible paralysis, prolonged coma, or even mortality. Consequently, \textit{the clinical paradigm gradually shifts to a \textbf{“diagnose first, treat later”} strategy, which entails the explicit prioritization of stereotactic biopsy (a navigation-assisted technique, please see \textbf{Sup. X})} to reliably obtain definitive pathological evidence before determining the optimal therapeutic regimen.

\begin{figure*}[t]
  \makebox[\textwidth][c]{%
    \includegraphics[width=\textwidth]{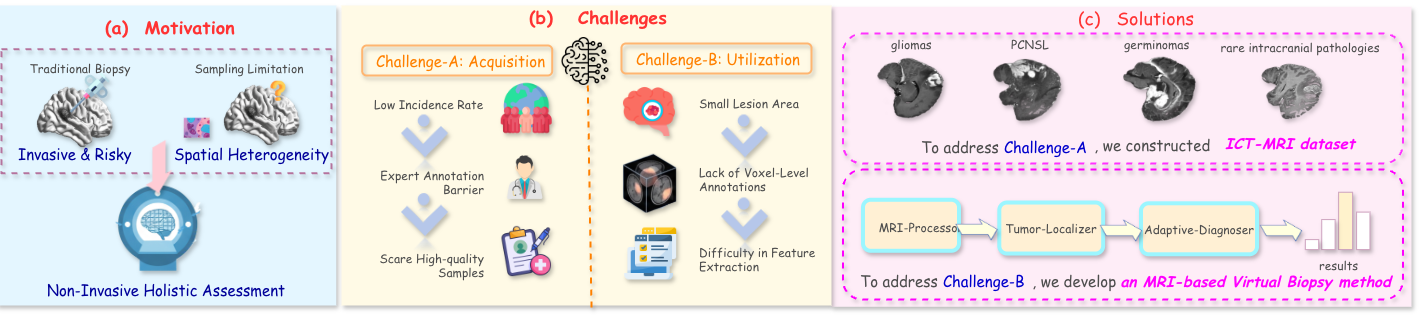}
  }
  \caption{%
  \textbf{(a) Motivation.} Traditional stereotactic biopsy for deep intracranial tumors carries inherent risks (hemorrhage, neurological deficits, tumor seeding) and suffers from sampling bias due to spatial heterogeneity. Non-invasive MRI-based pathology prediction is urgently needed for holistic tumor assessment.
  \textbf{(b) Challenges.} Challenge-A (Acquisition Difficulty): low incidence, long collection cycles, and expert annotation requiring biopsy-verified pathology; Challenge-B (Utilization Difficulty): tiny lesion volumes and lack of segmentation masks cause critical features to be overwhelmed by background noise.
  \textbf{(c) Solutions.} We construct the ICT-MRI dataset—the first public biopsy-verified benchmark with 249 cases across four categories; and develop a Virtual Biopsy framework with three modules (MRI-Processor, Tumor-Localizer, Adaptive-Diagnoser with MCA) achieving >90\% accuracy, outperforming baselines by over 20\%.
  }
  \label{fig:motivation}
\end{figure*}

\paragraph{Motivation.} 
However, stereotactic biopsy, despite being minimally invasive, carries inherent risks of hemorrhage, neurological deficits, and tumor seeding~\citep{field2001comprehensive}. 
Furthermore, biopsy is limited by tumor spatial heterogeneity (\textit{i.e.}, due to internal non-uniformity, a small, localized sample often fails to represent the whole mass); this restriction leads to sampling bias, false negatives, or inaccurate subtyping.
Therefore, there is an urgent clinical need for a non-invasive preoperative method to predict tumor pathology. Such an approach would substantially minimize unnecessary invasive procedures and naturally complement traditional biopsy by providing a holistic assessment, thereby effectively mitigating the inherent limitations of focal sampling.

Leveraging advances in computer vision algorithms and vision-language models (VLMs)~\citep{bordes2024introduction}, we aim to propose a Magnetic Resonance Imaging (MRI) based technique that deeply mines high-dimensional radiomic data, including subtle textures and morphological changes invisible to the naked eye, to non-invasively “decode” underlying biological attributes and support clinical decision-making.

\paragraph{Current Status and Challenges.} 

Current research on MRI-based intelligent diagnosis predominantly focuses on common tumors located in the brain periphery or non-functional areas. For example, these studies rely on public datasets, such as Kaggle~\citep{} and BraTS~\citep{menze2014multimodal}, and target recognition tasks for major categories like meningiomas and pituitary tumors. Conversely, research on deep intracranial tumors remains rarely explored despite their high diagnostic difficulty and surgical risk.
Conversely, research on deep intracranial tumors remains rarely explored despite their high diagnostic difficulty and surgical risk.

The primary bottleneck hindering advancement in this field lies in data, specifically manifesting as two distinct barriers:
(\texttt{Challenge-A}: Difficulty in Acquisition) The low incidence of deep tumors necessitates long collection cycles for raw data. More critically, annotation imposes a high barrier to entry; it requires high-level neurosurgical expertise to confirm diagnoses based on biopsy-proven pathology, making high-quality, biologically validated samples exceptionally scarce.
(\texttt{Challenge-B}: Difficulty in Utilization) Deep lesions occupy a minute fraction of the whole-brain field of view and are embedded within complex anatomical structures. Compounded by the lack of voxel-level fine annotations (segmentation masks), models lack explicit spatial guidance, causing critical lesion features to be easily overwhelmed by massive amounts of irrelevant background noise.

\paragraph{Solution.}

To address \texttt{Challenge-A}, we constructed a high-quality, dedicated benchmark termed the \textbf{I}ntra\textbf{C}ranial \textbf{T}umor MRI (\textbf{ICT-MRI}) dataset. This dataset comprises 249 patient samples spanning four distinct categories, including gliomas, primary central nervous system lymphoma (PCNSL), germinomas, and other rare intracranial pathologies. To guarantee data integrity and diagnostic authority, every case was rigorously verified via stereotactic biopsy with definitive histopathological confirmation. Regarding imaging modality, we standardized on T1-weighted contrast-enhanced (T1-CE) sequences. 
To our best knowledge, this is the first public dataset dedicated to intracranial tumors.





\begin{figure*}[t]
  \centering
  \includegraphics[width=\textwidth]{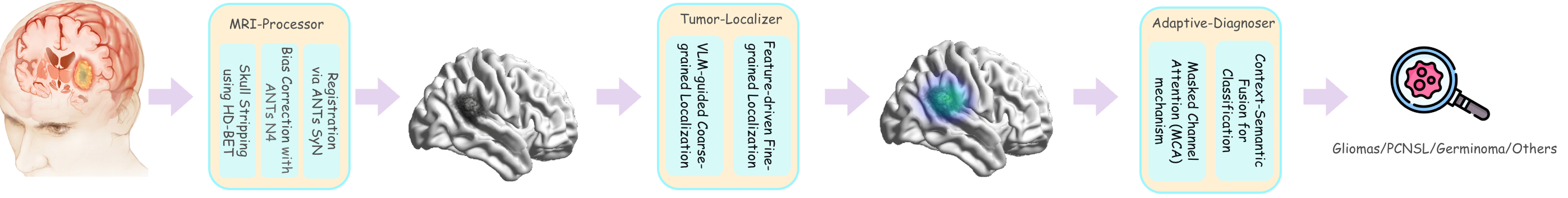}
  \caption{%
  Flowchart of the Whlole Module
}
  \label{fig:pipeline}
\end{figure*}

To address \texttt{Challenge-B}, we develop an MRI-based \textbf{Virtual Biopsy} (see Fig.~\ref{fig:pipeline}) method integrating computer vision and VLM technologies. 
The proposed architecture consists of three core components: 
(i) The MRI Pre-processing Module (\textbf{MRI-Processor}, see Sup.~X) employs mature tools~\citep{isensee2019automated}, \citep{avants2009advanced} to standardize raw MRI scans, thereby mitigating data inconsistency and establishing a solid foundation for robust, generalizable modeling.
(ii) The Intracranial Tumor Localization Module (\textbf{Tumor-Localizer}, see Fig.~\ref{fig:localization}) implements a coarse-to-fine strategy; it initially leverages VLMs to generate weak voxel-level supervision for coarse localization, subsequently employing a lightweight model to achieve feature-driven fine-grained localization. 
(iii) The Adaptive Diagnostic Module (\textbf{Adaptive-Diagnoser}, see Fig.~\ref{fig:Diagnoser}) incorporates a novel Masked Channel Attention (MCA) mechanism to highlight discriminative regions, fusing local attention-refined features with global whole-brain contexts to yield a precise diagnosis.

\paragraph{Contribution.}

(i) We release the ICT-MRI Dataset, the first biopsy-verified benchmark dedicated to deep intracranial tumors, effectively addressing the field's critical data scarcity.
(ii) We propose a VLM-driven Virtual Biopsy framework. By mining invisible radiomic features, it non-invasively decodes pathological attributes to complement traditional biopsy.
(iii) Experiments show our method achieves $>$ 90\% accuracy, outperforming other related approaches by over 20\% and demonstrating strong potential for clinical deployment.

\section{Related Work}


\paragraph{Tumor Recognition.} Early approaches to automated brain tumor diagnosis primarily relied on handcrafted features extracted from MRI, such as texture and intensity histograms, combined with traditional classifiers like support vector machines (SVMs) ~\citep{avants2009advanced}~\citep{vani2017brain}~\citep{nandpuru2014mri}. However, these methods suffered from limited generalization and high sensitivity to noise. With the advent of deep learning, convolutional neural networks (CNNs) have emerged as the dominant paradigm. In particular, U-Net and its variants (e.g., nnU-Net~\citep{isensee2020nnu}~\citep{isensee2021nnu}) leverage encoder-decoder architectures to effectively capture local contextual information, establishing strong benchmarks for brain tumor diagnosis. Despite their strength in feature representation, CNNs are constrained by a limited receptive field and struggle to model long-range dependencies. To address this, Transformer-based architectures (e.g., Swin UNETR~\citep{tang2022self}) have been introduced, employing self-attention mechanisms to capture global semantic information, thereby enhancing the recognition of complex tumors. Meanwhile, foundation models such as the Segment Anything Model (SAM)~\citep{ali2024evaluating} and its medical adaptations (MedSAM)~\citep{zhou2025medsam} offer new possibilities for zero-shot generalization across different scanning protocols and tumor types.

In recent years, significant progress has been made in brain tumor classification~\citep{paul2017deep}. One line of research focuses on developing high-precision classifiers to distinguish between different tumor types, thereby improving clinical decision-making efficiency; these approaches typically rely on large-scale clinical datasets to approach expert-level diagnostic accuracy~\citep{talukder2025deep}. Another line emphasizes integrating multi-class segmentation and classification within a unified framework~\citep{lu2025deep}. By leveraging deep convolutional networks or Transformer-based fusion mechanisms, these methods enable fine-grained extraction of tumor cores, peritumoral edema, and enhancing tumor regions, providing critical spatial information for subsequent grading and treatment planning~\citep{karagoz2024residual}~\citep{benzorgat2024enhancing}. Furthermore, to address data scarcity and domain shifts caused by heterogeneous scanning protocols, ensemble learning and self-supervised pretraining strategies have been adopted, significantly improving model robustness and generalization across diverse imaging conditions~\citep{ge2024novel}~\citep{fang2021self}.
\paragraph{Large Models for Semantic Reasoning and Annotation.}
Although traditional diagnostic models focus on pixel-level predictions, medical image interpretation often requires higher-level semantic reasoning. The rapid development of large language models (LLMs) has given rise to large visual-language models (LVLMs), enabling the integration of visual understanding with natural language reasoning for tasks such as diagnostic report generation and biomedical question answering. Early adaptation efforts, such as LLaVA-Med~\citep{li2023llava}, align visual features with language embeddings, providing preliminary medical image understanding, but they remain limited in capturing fine-grained visual details, such as small lesion detection. To overcome these limitations, the Qwen-VL~\citep{bai2023qwen}~\citep{wang2024qwen2} series incorporates high-resolution visual inputs and advanced vision-language alignment techniques, offering strong grounding and OCR capabilities to support dense descriptions and automated annotation. Notably, the latest Qwen3-VL~\citep{qwen3vl2025} demonstrates significant advances in visual reasoning and document understanding, enabling zero-shot recognition of complex medical structures and generation of structured descriptive labels. Leveraging Qwen3-VL for automated or semi-automated annotation not only alleviates the high labor cost of manual labeling but also maintains high diagnostic accuracy. Building on this trajectory, our study integrates Qwen3-VL as the core annotation engine, bridging high-level semantic reasoning with low-level tumor feature extraction, thereby addressing the challenge of obtaining high-quality annotations in medical scenarios.
\section{Data Collection}

We retrospectively collected a cohort of patients with deep intracranial tumors treated at [Institution Name] between [Start Year] and [End Year]. To ensure data authority and label reliability, strict inclusion criteria were established: (1) patients underwent standard preoperative T1-weighted contrast-enhanced (T1-CE) MRI scanning; and (2) all lesions had definitive histopathological diagnoses confirmed via stereotactic biopsy. Consequently, we constructed a high-quality ICT-MRI dataset comprising 249 patients. This study was approved by the Institutional Review Board (IRB), and all patient sensitive information was strictly de-identified.Regarding demographic characteristics, the cohort includes 135 males and 114 females. The patient ages ranged from 4 to 86 years, with a mean age of 43.7 $\pm$ 19.7 years. In terms of pathological distribution, the dataset covers four representative categories of deep intracranial tumors. Detailed demographic statistics and pathological distributions are provided in Tab.~\ref{tab:Patient_Cohort}.

\begin{table}[htbp]
    \centering
    \includegraphics[width=1.0\linewidth]{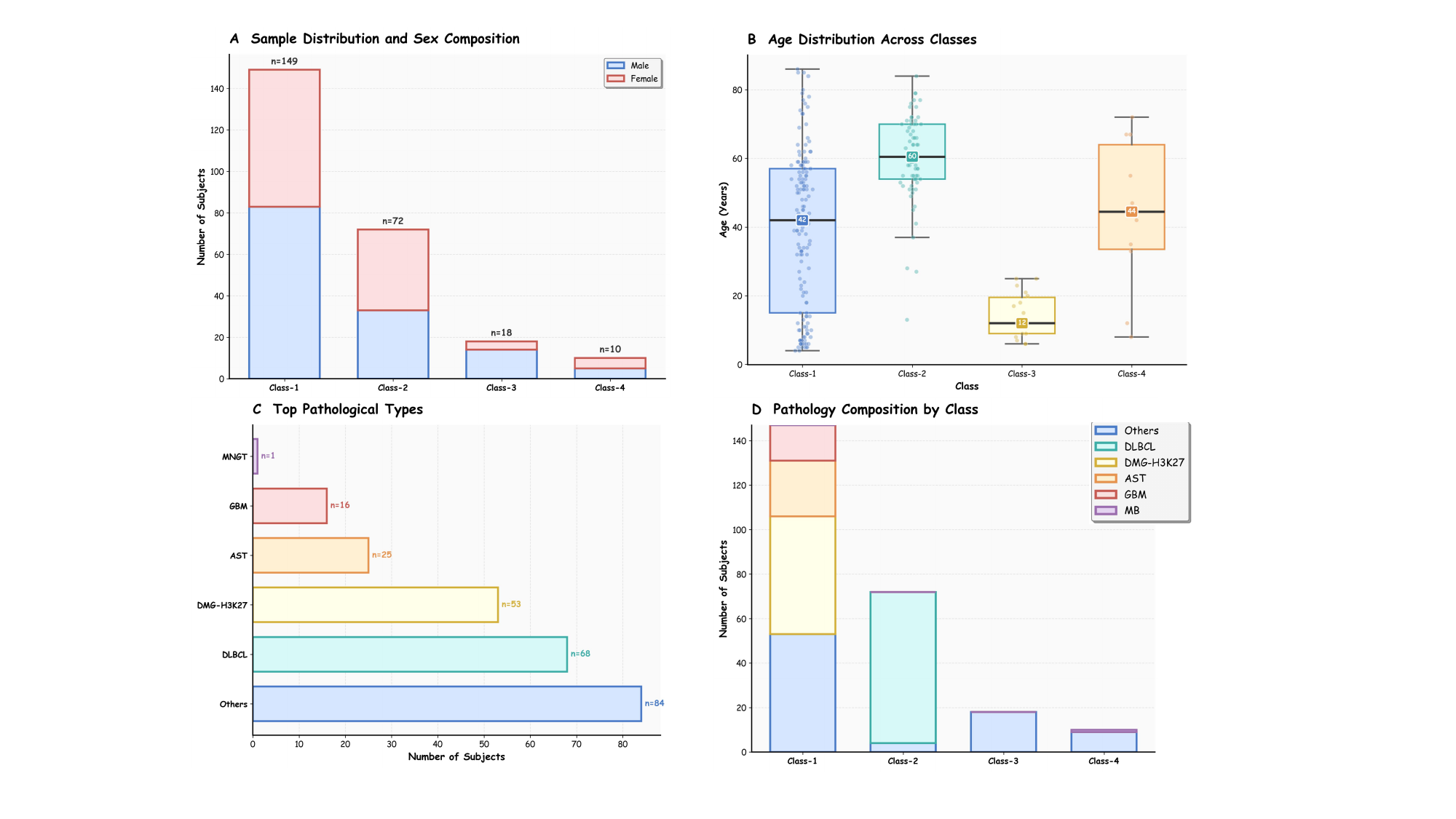}
    \caption{Demographic and Pathological Characteristics of the Patient Cohort. Abbreviations: GBM, glioblastoma; AST, astrocytoma; ODG, oligodendroglioma; DMG-H3K27, diffuse midline glioma with H3 K27 alteration; DLBCL, diffuse large B-cell lymphoma; EPN, ependymoma; MB, medulloblastoma; MNGT, mixed neuronal--glial tumor.}
    \label{tab:Patient_Cohort}
\end{table}

\section{Methodology}

\subsection{Overview}

Our proposed method consists of three key modules (see Fig.~\ref{fig:pipeline}): the MRI Pre-processing Module (\textbf{MRI-Processor}), the Intracranial Tumor Localization Module (\textbf{Tumor-Localizer}), and the Adaptive Diagnostic Module (\textbf{Adaptive-Diagnoser}).

\begin{figure*}[t]
  \centering
  \includegraphics[width=\textwidth]{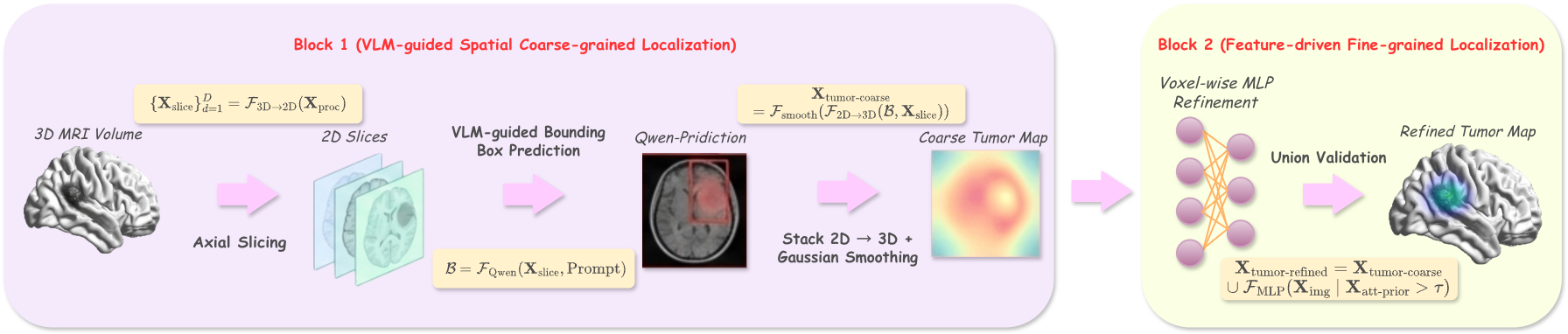}
  \caption{%
  The flowchart of Intracranial Tumor Localization Module (Tumor-Localizer).
  Block 1 (VLM-guided Spatial Coarse-grained Localization) consists of three sequential steps: (i) Perform axial slicing to convert the preprocessed 3D MRI into a sequence of 2D cross-sectional images; (ii) Feed each 2D slice along with structured radiological prompts to the Vision-Language Model (Qwen3-VL) to predict bounding boxes based on abnormal signal patterns and anatomical asymmetry; (iii) Spatially stack all bounding boxes and apply Gaussian smoothing with morphological dilation to generate a smooth 3D spatial prior map, serving as initial coarse-grained attention. Block 2 (Feature-driven Fine-grained Localization) comprises two steps: (i) Utilize the coarse spatial prior as pseudo-labels to train a lightweight voxel-wise MLP classifier on high-confidence regions, learning intrinsic tumor-background discriminative patterns; (ii) Perform union validation between MLP predictions and VLM-based results to maximize recall and produce the final refined spatial prior map that accurately captures irregular tumor boundaries.
}
  \label{fig:localization}
\end{figure*}
\subsection{MRI Pre-processing Module}

Raw 3D MRI data typically exhibit large variability across scanners and subjects, including differences in spatial resolution, contrast, and signal-to-noise ratio. To ensure robustness and improve cross-subject comparability, we adopt a standardized three-stage MRI pre-processing pipeline applied to all T1-weighted images. Specifically, the pipeline consists of: (i) skull stripping using HD-BET ~\citep{isensee2019automated} to remove non-brain tissue; (ii) bias field correction with ANTs N4 ~\citep{avants2009advanced} to mitigate intensity inhomogeneity; and (iii) nonlinear registration via ANTs SyN ~\citep{avants2009advanced} to align individual scans to a standardized anatomical template (MNI space).
Let the raw MRI volume be denoted as $\mathbf{X}_{\text{raw}} \in \mathbb{R}^{D \times H \times W}$. The workflow can be formulated as:
\begin{equation}
\mathbf{X}_{\text{proc}} = \mathcal{F}_{\text{SyN}}\!\left(
\mathcal{F}_{\text{N4}}\!\left(
\mathcal{F}_{\text{BET}}\!\left(\mathbf{X}_{\text{raw}}\right)
\right)\right),
\label{eqa: pre_processing}
\end{equation}
where $\mathbf{X}_{\text{proc}} \in \mathbb{R}^{D \times H \times W}$ denotes the processed MRI volume. Here, $\mathcal{F}_{\text{BET}}$, $\mathcal{F}_{\text{N4}}$, and $\mathcal{F}_{\text{SyN}}$ represent the operations of brain extraction, bias correction, and nonlinear spatial normalization, respectively, and $D$, $H$, and $W$ correspond to spatial dimensions (depth, height, and width).

\subsection{Intracranial Tumor Localization Module}
The \textbf{Tumor-Localizer} (see Fig.~\ref{fig:localization}) is designed to address \texttt{Challenge-B}, it effectively prevents critical information from being overwhelmed by background noise. By accurately localizing and extracting discriminative features, it provides refined and high-value feature inputs for the subsequent diagnosis. It includes two main blocks:

\paragraph{Block 1: VLM-guided Spatial Coarse-grained Localization.}


Due to the lack of voxel-level annotations in the collected data, we leverage the powerful multimodal reasoning capabilities of a VLM (specifically, Qwen3-VL~\citep{qwen3vl2025}) to perform coarse localization of intracranial tumors.

First, considering that the model cannot directly process 3D volumetric data, we perform axial slicing on the pre-processed 3D MRI scans to convert them into a sequence of 2D cross-sectional images, which can be formulated as:
\begin{equation}
\{\mathbf{X}_{\text{slice}}\}_{d=1}^{D} = \mathcal{F}_{\text{3D}\to\text{2D}}\!\left(
\mathbf{X}_{\text{proc}}\right),
\label{eqa: slice_3D-2D}
\end{equation}
where 
$\mathcal{F}_{\text{3D}\to\text{2D}}$ denotes the 3D-to-2D axial slicing operation;
$\mathbf{X}_{\text{slice}} \in \mathbb{R}^{H \times W}$ is $d$-th slice.

Subsequently, we employ expert-driven prompts (\textit{e.g.}, abnormal signal intensities, irregular boundaries, edema-associated gradients, and midline asymmetry) to guide the model in assessing tumor likelihood and generating predicted bounding boxes on 2D slices,  which can be formulated as:
\begin{equation}
\mathcal{B} = \mathcal{F}_{\text{Qwen}}\!\left(
\mathbf{X}_{\text{slice}}, \text{Prompt}
\right),
\label{eqa: bounding_box}
\end{equation}
where 
$\mathcal{F}_{\text{VLM}}$ denotes the operation using Qwen model;
$\mathcal{B}$ is the predicted bounding box on 2D slice.

Next, we spatially stack all 2D slices containing predicted bounding boxes to preliminarily reconstruct a 3D candidate region. However, the 3D volume generated by directly stacking 2D bounding boxes exhibits spatial discreteness and discontinuity, often manifesting as stair-step artifacts. To address this and mitigate the risk of incomplete tumor coverage due to potential VLM prediction biases, we apply Gaussian smoothing and morphological dilation. This post-processing step transforms the discrete bounding boxes into a continuous probability distribution, yielding a smooth 3D spatial prior map. \textit{This map serves as an initial coarse-grained attention, providing global guidance for the subsequent fine-grained location refinement,} which can be expressed as:
\begin{equation}
\mathbf{X}_{\text{tumor-coarse}} = \mathcal{F}_{\text{smooth}} \!\left(
\mathcal{F}_{\text{2D}\to\text{3D}}\!\left(
\mathcal{B}, \mathbf{X}_{\text{slice}}
\right)\right),
\label{eqa: coarse_tumor}
\end{equation}
where $\mathcal{F}_{\text{2D}\to\text{3D}}$ and  $\mathcal{F}_{\text{smooth}}$ denote the 2D-to-3D stacking operation and the Gaussian smoothing operation, respectively.
$\mathbf{X}_{\text{tumor-coarse}} \in \mathbb{R}^{D \times H \times W}$ represents the coarse-grained tumor spatial.
Please see \textbf{Sup. I} for more details.

\paragraph{Block 2: Feature-driven Fine-grained Localization.}

Next, leveraging the coarse tumor spatial priors generated by Eq.~(\ref{eqa: coarse_tumor}), we obtain weakly supervised voxel-level labels and initiate a self-correction process. \textit{This step is necessary because probability maps derived directly from bounding boxes are spatially overly smooth, failing to capture irregular boundaries and inevitably including non-tumor background tissue.}
To address this, we utilize the coarse voxel-level annotation as pseudo-labels to train a lightweight voxel-wise Multi-Layer Perceptron (MLP). This MLP is tasked with learning the intrinsic visual consistency of the tumor pixels. We then validate the MLP predictions against the VLM-based results by taking the union of both, aiming to maximize the recall and encompass all voxels potentially involved in the tumor. The process can be formulated as:
\begin{equation}
\mathbf{X}_{\text{tumor-refined}} = \mathbf{X}_{\text{tumor-coarse}} \cup \mathcal{F}_{\text{MLP}}(\mathbf{X}_{\text{img}} | \mathbf{X}_{\text{att-prior}} > \tau),
\label{eqa: refined_tumor}
\end{equation}
where 
$\mathcal{F}_{\text{MLP}}$ denotes the voxel-wise prediction of the MLP trained on high-confidence regions (thresholded by $\tau$); 
$\cup$ represents the pixel-wise union operation; 
$|$ denotes a conditional indexing operation, indicating that the MLP is trained explicitly on voxels where the initial VLM-based confidence exceeds;
and $\mathbf{X}_{\text{tumor-refined}}$ is the corrected spatial prior map.

\subsection{Adaptive Diagnostic Module}

\begin{figure}[htbp]
    \centering
    \includegraphics[width=1.0\linewidth]{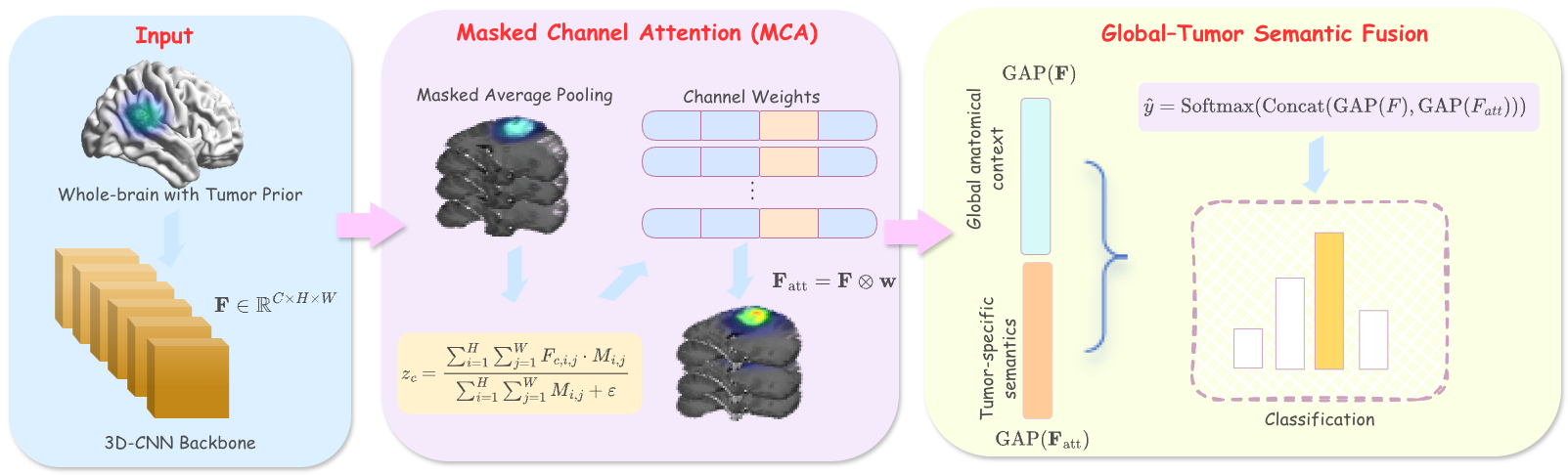}
    \caption{The flowchart of Adaptive Diagnostic Module (Adaptive-Diagnoser). The module comprises three sequential components: (i) Input: Extract intermediate feature maps from the preprocessed whole-brain MRI using a 3D-CNN backbone, while obtaining the refined spatial mask from the Tumor-Localizer; (ii) Masked Channel Attention (MCA): Perform masked average pooling exclusively on tumor regions to compute channel-wise statistics that extract pure tumor semantics without background interference, then generate adaptive channel weights through a MLP and recalibrate the feature maps via channel-wise multiplication to enhance tumor-specific discriminative channels while suppressing background noise; (iii) Global–Tumor Semantic Fusion: Aggregate global context features and refined tumor-specific features through Global Average Pooling, and produce the final probability distribution.}
    \label{fig:Diagnoser}
\end{figure}

The \textbf{Adaptive-Diagnoser} (see Fig.~\ref{fig:Diagnoser}) integrates the attention mechanism to yield the final diagnosis.


Guided by clinical prior knowledge, we recognize that the primary distinction among intracranial tumor subtypes lies in their varying intratumoral heterogeneity and signal intensity patterns. To capture these case-specific nuances, we introduce the Masked Channel Attention (MCA) mechanism to achieve adaptive feature refinement.
\textit{This design is motivated by the nature of deep feature representations, where distinct channels encode specific semantic attributes, such as texture regularity, gradient patterns, or tissue-specific responses. Consequently, the MCA mechanism assigns adaptive importance weights to recalibrate these channels. This process automatically focuses the model on tumor-specific discriminative traits while suppressing irrelevant background noise.}
Formally, we define the input feature map $\mathbf{F}$ and the corresponding spatial mask $\mathbf{M}$ as:
\begin{equation}
\mathbf{F} = \mathcal{F}_{\text{3D-CNN}} \left(
\mathbf{X}_{\text{proc}}
\right), \quad \mathbf{M} = \mathbf{X}_{\text{tumor-refined}},
\label{eqa: attention_input}
\end{equation}
where $\mathbf{F} \in \mathbb{R}^{C \times H \times W}$ represents the intermediate feature maps extracted by the backbone, and $\mathbf{M} \in \mathbb{R}^{1 \times H \times W}$ denotes the refined spatial prior obtained from the previous stage.
In deep feature maps, distinct channels encode specific semantic attributes (such as texture regularity, gradient patterns, or tissue-specific responses). Consequently, our mechanism assigns adaptive importance weights to different channels. This process automatically focuses the model on channels that capture tumor-specific discriminative traits while suppressing those activated by irrelevant background noise.
\textit{Notably, using whole-brain input effectively retains critical global context, such as symmetry. Unlike hard cropping, which risks losing peripheral details, this approach enables “soft” attentional filtering, allowing the model to better contrast tumor features against the background.}

To extract pure tumor semantics without background interference, we perform Masked Average Pooling to compute the channel-wise statistic $\mathbf{z} \in \mathbb{R}^{C}$:
\begin{equation}
z_c = \frac{\sum_{i=1}^{H} \sum_{j=1}^{W} \mathbf{F}_{c,i,j} \cdot \mathbf{M}_{i,j}}{\sum_{i=1}^{H} \sum_{j=1}^{W} \mathbf{M}_{i,j} + \epsilon},
\label{eqa: masked_pooling}
\end{equation}
where $z_c$ is the descriptor for the $c$-th channel, and $\epsilon$ is a small constant for numerical stability. Finally, the adaptive channel weights $\mathbf{w}$ are generated to recalibrate the feature maps:
\begin{equation}
\mathbf{w} = \sigma\left(\mathcal{F}_{\text{MLP}}(\mathbf{z})\right), \quad \mathbf{F}_{\text{att}} = \mathbf{F} \otimes \mathbf{w},
\label{eqa: recalibration}
\end{equation}
where $\mathcal{F}_{\text{MLP}}$ consists of two fully connected layers with a ReLU activation in between to capture channel dependencies; $\sigma$ denotes the Sigmoid function; and $\otimes$ represents channel-wise multiplication.

Subsequently, we aggregate the global context with the tumor-specific semantics for the final classification. This strategy aims to preserve essential global structural references while focusing on discriminative tumor traits. Formally, this process can be formulated as:
\begin{equation}
    \hat{\mathbf{y}} = \text{Softmax}
    \left(
    \text{Concat}
    \left(
    \text{GAP}(\mathbf{F}),\text{GAP}(\mathbf{F}_{\text{att}})
    \right) \right)
    \label{eqa: classification}
\end{equation}
where $\hat{\mathbf{y}} \in \mathbb{R}^{K}$ denotes the predicted probability distribution over $K$ tumor classes. $\text{GAP}(\cdot)$ represents the Global Average Pooling operation, which spatially aggregates the feature maps into compact vectors. $\text{Concat}(\cdot)$ refers to the concatenation operation along the channel dimension, fusing the global context and refined semantics. Note that a linear classifier is implicitly applied to the concatenated features before Softmax.
We use the Cross-Entropy Loss to update the network.

\begin{figure}[htbp]
    \centering
    \includegraphics[width=1.0\linewidth]{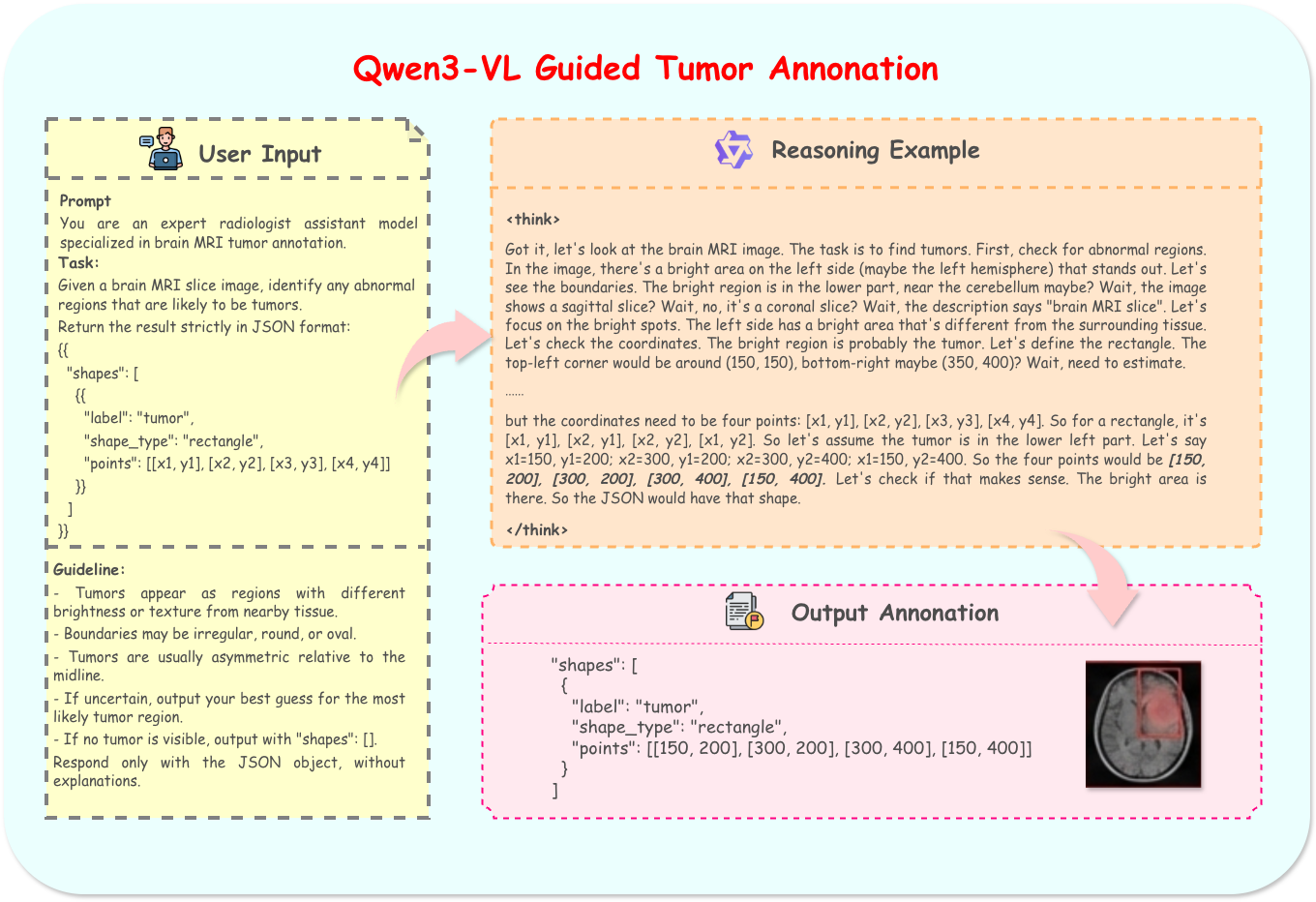}
    \caption{Brain MRI Tumor Annonation }
    \label{fig:vlm_prompt}
\end{figure}

\newpage

\section{Experiments}
\textbf{Data.}
We constructed a high-quality ICT-MRI dataset comprising 249 patients across four tumor categories, as summarized in Tab.~\ref{tab:Patient_Cohort}.
Following standard practice in medical imaging, we use stratified 5-fold patient-level cross-validation to ensure robust evaluation and prevent data leakage. In each fold, 80\% of patients are used for training and 20\% for testing, with no overlap between subsets. Reported results are averaged across folds to ensure statistical reliability.

\textbf{Data Augmentation.} We apply spatial and intensity augmentations to improve 
generalization. Spatial augmentations include random 90° rotations along the axial 
plane (probability 0.5) and random flipping along the sagittal axis (probability 0.5). 
Intensity augmentations include Gaussian noise injection ($\sigma=0.01$, probability 0.15) 
and contrast adjustment via gamma correction ($\gamma \in [0.9, 1.1]$, probability 0.15). 
Augmentations are applied consistently to both the full volume and ROI to maintain spatial correspondence. 

\textbf{Metrics.} We evaluate classification performance using accuracy, precision, recall, and F1-score. While accuracy provides an overall measure of prediction correctness, precision reflects the model's ability to avoid false positives, and recall captures sensitivity to true positives—an especially critical aspect in medical imaging, where missed detections may have clinical consequences. The F1-score serves as a balanced harmonic measure of precision and recall, making it more robust under class imbalance. Together, these metrics offer a comprehensive and reliable assessment of model performance beyond accuracy alone.

\textbf{Implementation.} Our method is implemented in Pytorch and  trained on NVIDIA RTX 4090 GPUs for up to 250 epochs   using the AdamW optimizer with a learning rate of $5\times10^{-5}$ and a weight decay of $1\times10^{-4}$. The learning rate was scheduled using cosine annealing with warm restarts(initial period $T_0=10$ epochs). Gradient clipping (max norm 2.0) and dropout ($p=0.2$) were applied for stability. Label smoothing ($\epsilon=0.05$) was used to reduce overconfident predictions. To simulate an effective batch size of 4 while preserving memory efficiency, training adopted gradient accumulation over two steps with a batch size of 2. 

\begin{figure}[htbp]
    \centering
    \includegraphics[width=1.0\linewidth]{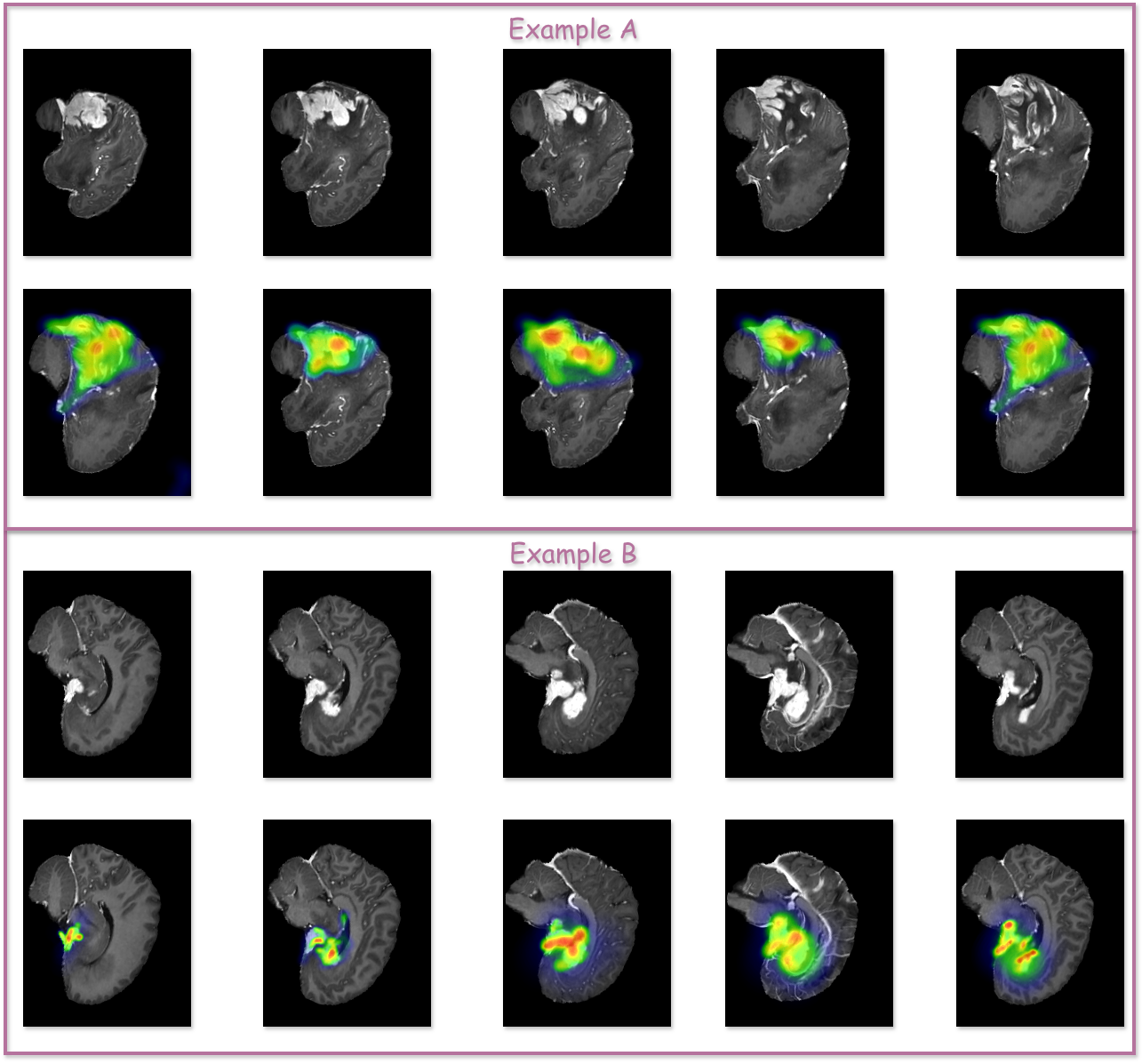}
    \caption{Brain MRI Tumor Annonation }
    \label{fig:gradcam}
\end{figure}

\textbf{Interpretability Validation of Virtual Biopsy Framework.}To validate the clinical rationality of our framework, we employ 3D Grad-CAM~\citep{selvaraju2017grad} to visualize the spatial activation patterns of the complete model, encompassing both the Tumor-Localizer and the MCA attention mechanism. We randomly select two representative cases from the test set and present their volumetric heatmaps through axial slice visualization, as shown in Fig.~\ref{fig:gradcam}.

We invite a panel of experienced neuroradiologists, including senior neurosurgeons (Associate Chief Physician level or above) and attending radiologists with extensive experience in intracranial tumor interpretation, to independently evaluate the visualization results. The clinical assessment reveals two key findings: (1) \textit{Model attention aligns with diagnostic reasoning.} The high-response regions in the heatmaps precisely coincide with tumor-related structures, and the model's attention focus demonstrates strong agreement with the critical areas that clinicians prioritize during diagnostic evaluation. This indicates that our MCA mechanism effectively captures discriminative features essential for tumor classification. (2) \textit{VLM-guided priors exhibit clinical plausibility.} Experts confirm that the regions highlighted by the model's attention mechanism accurately correspond to the lesion boundaries and surrounding edema zones, which validates the effectiveness of the VLM-guided spatial prior ($\mathbf{X}_{\text{tumor-coarse}}$ in Eq.~(\ref{eqa: coarse_tumor})) in providing reliable initial localization guidance.

These results collectively demonstrate that our framework—from VLM-based coarse localization to MCA-driven fine-grained attention—exhibits clinical interpretability and trustworthiness throughout the entire diagnostic pipeline, thereby enhancing both model transparency and practical applicability in real-world clinical scenarios.

\begin{figure*}[t]
  \centering
  \includegraphics[width=\textwidth]{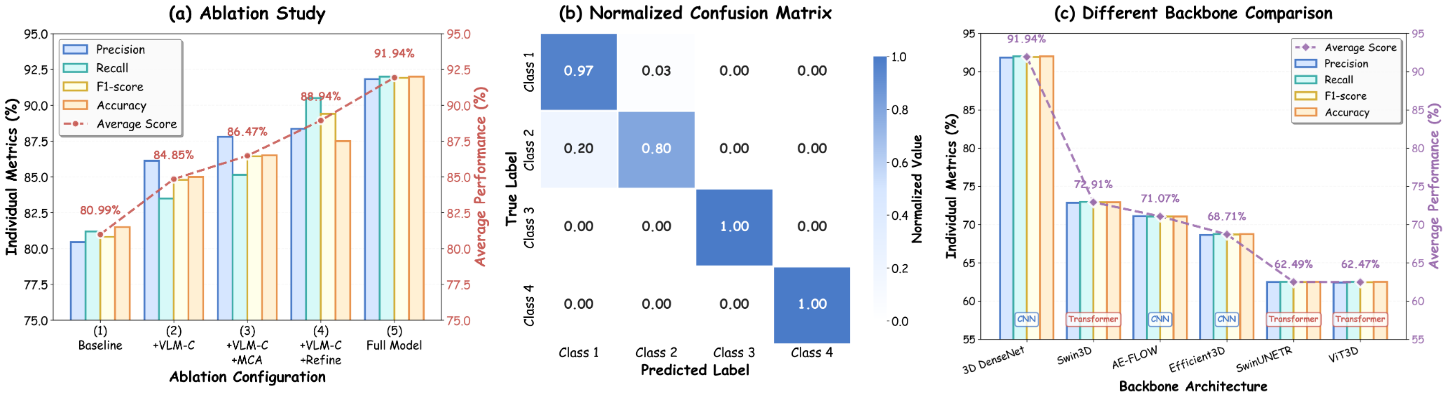}
  \caption{Comprehensive evaluation of the proposed model. (a) Ablation study evaluates the contribution of each module. (b) Normalized confusion matrix shows classification performance across classes. (c) Backbone comparison presents performance differences between CNN- and Transformer-based architectures.}
    \label{fig:overall_evaluation}
\end{figure*}
\subsection{MRI Pre-processing Module}

\textbf{Ablation Studies.}  
Our method comprises the \textit{Intracranial Tumor Localization Module (Loc.)} and the \textit{Adaptive Diagnostic Module (Diag.)}. We assess each module via ablation; detailed results appear in Tab.~\ref{tab:ablation}. These results are further illustrated in Fig.~\ref{fig:overall_evaluation}(a), which visualizes Precision, Recall, F1-score, Accuracy, and the Average Score across different module configurations.

For Tumor-Localizer. (i) Row \textcircled{1} serves as the baseline, supervising the classifier directly on raw MRI volumes without any localization. Compared with Row \textcircled{5}, all metrics drop sharply (e.g., Accuracy decreases by 10.5\%), indicating that the vast background noise in 3D MRI overwhelms the critical tumor features, underscoring the necessity of the Localization Module. (ii) Row \textcircled{2} introduces the VLM-guided coarse localization. While Precision improves significantly over the baseline due to background removal, the Recall (83.50\%) remains suboptimal. This suggests that the "stair-step" artifacts from stacked 2D boxes may miss peripheral tumor regions. (iii) Row \textcircled{4} incorporates the Refine-MLP. Compared with Row \textcircled{2}, the Recall increases dramatically by 7.0\% (from 83.50\% to 90.50\%), while Accuracy reaches 87.50\%. This confirms that the union validation mechanism in Eq.~(\ref{eqa: refined_tumor}) effectively recovers missed tumor voxels and corrects the spatial discontinuity of VLM predictions.

For Adaptive-Diagnoser. (i) Row \textcircled{3} applies the MCA module directly on the coarse VLM regions (without Refine). Compared with Row \textcircled{2}, the F1-score improves by approximately 1.6\%, demonstrating that feature recalibration is beneficial even with coarse spatial priors. However, it still lags behind the full model, as the attention mechanism is sensitive to the quality of the spatial mask. (ii) Row \textcircled{5} integrates MCA with the refined localization (Full Model). Compared with Row \textcircled{4} (Loc. only), Accuracy and F1-score increase by 4.5\% and 2.5\% respectively, achieving the best performance of \textbf{92.00\%}. This demonstrates that MCA successfully captures the intratumoral heterogeneity by adaptively weighting channels, but its full potential is unlocked only when coupled with precise spatial refinement.
To further illustrate the classification capability of our full model, the class-wise prediction performance is visualized via the normalized confusion matrix in Fig.~\ref{fig:overall_evaluation}(b), highlighting the model's ability to correctly classify different tumor types.

\begin{table}[t]
    \centering
    \caption{Ablation studies on the effectiveness of proposed modules. \textit{VLM-C}: VLM-guided Spatial Coarse-grained Localization; \textit{Refine}: Feature-driven Fine-grained Localization; \textit{MCA}: Masked Channel Attention.}
    \label{tab:ablation}
    \resizebox{0.48\textwidth}{!}{%
    \begin{tabular}{lccccccc}
        \toprule
        \multirow{2}{*}{\textbf{ID}} & \multicolumn{2}{c}{\textbf{Tumor-Loc.}} & \textbf{Adaptive-Diag.} & \multicolumn{4}{c}{\textbf{Metrics (\%)}} \\
        \cmidrule(lr){2-3} \cmidrule(lr){4-4} \cmidrule(lr){5-8}
         & VLM-C & Refine & MCA & Precision & Recall & F1-score & Accuracy \\
        \midrule
        \textcircled{1} &  &  &  & 80.45 & 81.20 & 80.82 & 81.50 \\
        \textcircled{2} & \checkmark &  &  & 86.12 & 83.50 & 84.79 & 85.00 \\
        \textcircled{3} & \checkmark &  & \checkmark & 87.80 & 85.15 & 86.45 & 86.50 \\
        \textcircled{4} & \checkmark & \checkmark &  & 88.35 & 90.50 & 89.41 & 87.50 \\
        \textcircled{5} & \checkmark & \checkmark & \checkmark & \textbf{91.83} & \textbf{92.00} & \textbf{91.91} & \textbf{92.00} \\
        \bottomrule
    \end{tabular}%
    }
\end{table}

\textbf{Different Backbone.}  
Our framework is \textit{model-agnostic} and compatible with various backbone architectures. Unless otherwise specified, all experimental results are reported using 3D DenseNet~\citep{huang2017densely} as the default backbone.

To validate this choice, we evaluate multiple representative architectures, including CNN-based models (AE-FLOW~\citep{zhao2023ae}, Efficient3D~\citep{kopuklu2019resource}) and Transformer-based models (ViT3D~\citep{chen2023masked}, Swin3D~\citep{chen2023masked}, SwinUNETR~\citep{tang2022self}). As shown in Table~\ref{tab:backbone_comparison}, \textit{3D DenseNet achieves the highest accuracy of 92.00\%}, significantly outperforming alternative backbones. Among the remaining methods, Swin3D demonstrates the best performance with an accuracy of 72.92\%, while AE-FLOW and Efficient3D achieve 71.05\% and 68.75\%, respectively. ViT3D and SwinUNETR exhibit limited effectiveness, each reaching an accuracy of 62.50\%, suggesting that vanilla vision transformer architectures require substantial task-specific adaptations for this problem. Overall, the superior performance of 3D DenseNet validates that its dense connectivity patterns are particularly effective for capturing the pathological variations of intracranial tumors.
Furthermore, the impact of different backbone architectures is summarized in Fig.~\ref{fig:overall_evaluation}(c).

\begin{table}[t]
\centering
\caption{Comparison (\%) of different backbones. Most results are reported with the 3D DenseNet. ViT3D, Swin3D, and SwinUNETR are based on the Transformer architecture, AE-FLOW, Efficient3D and 3D DenseNet are based on the CNN architecture.}
\label{tab:backbone_comparison}
\resizebox{\columnwidth}{!}{%
\begin{tabular}{lcccc}
\toprule
\textbf{Backbone} & \textbf{Precision(\%)} & \textbf{Recall(\%)} & \textbf{F1-score(\%)} & \textbf{Accuracy(\%)} \\
\midrule
3D DenseNet~\citep{huang2017densely} & \textbf{91.83} & \textbf{92.00} & \textbf{91.91} & \textbf{92.00} \\
Swin3D~\citep{chen2023masked}      & 72.85 & 72.97 & 72.91 & 72.92 \\
AE-FLOW~\citep{zhao2023ae}        & 71.10 & 71.05 & 71.07 & 71.05 \\
Efficient3D~\citep{kopuklu2019resource} & 68.65 & 68.75 & 68.70 & 68.75 \\
SwinUNETR~\citep{tang2022self}    & 62.48 & 62.50 & 62.49 & 62.50 \\
ViT3D~\citep{chen2023masked}      & 62.42 & 62.50 & 62.46 & 62.50 \\
\bottomrule
\end{tabular}%
}
\end{table}

\newpage

\bibliographystyle{named}
\bibliography{ijcai26}

\end{document}